\documentclass[letterpaper, 10 pt, conference]{ieeeconf}  % Comment this line out if you need a4paper
\usepackage{amsmath}
\usepackage{algpseudocode} 
\usepackage{graphicx}
\usepackage{amssymb}
\usepackage{algorithm}
\usepackage{xcolor}

\IEEEoverridecommandlockouts                              % This command is only needed if 
\title{\LARGE \bf
Learning to Pour
}

\author{Yongqiang Huang and Yu Sun% <-this % stops a space
\thanks{The authors are with the Department of Computer Science and Engineering, University of South Florida, Tampa, FL 33620, USA. Email: \texttt{yongqiang@mail.usf.edu, yusun@cse.usf.edu}}%
}

\begin{document}

\maketitle
\thispagestyle{empty}
\pagestyle{empty}

%%%%%%%%%%%%%%%%%%%%%%%%%%%%%%%%%%%%%%%%%%%%%%%%%%%%%%%%%%%%%%%%%%%%%%%%%%%%%%%%
\begin{abstract}
Pouring is a simple task people perform daily. It is the second most frequently executed motion in cooking scenarios, after pick-and-place. We present a pouring trajectory generation approach, which uses force feedback from the cup to determine the future velocity of pouring. The approach uses recurrent neural networks as its building blocks. We collected the pouring demonstrations which we used for training. To test our approach in simulation, we also created and trained a force estimation system. The simulated experiments show that the system is able to generalize to single unseen element of the pouring characteristics.         
\end{abstract}

%%%%%%%%%%%%%%%%%%%%%%%%%%%%%%%%%%%%%%%%%%%%%%%%%%%%%%%%%%%%%%%%%%%%%%%%%%%%%%%%
\section{Introduction}

Tasks that are found in manufacturing facilities are often precisely defined, repetitive, and tolerates very little error. Those tasks can be easily handled by industrial robots but proves difficult for human workers. In contrast, each time a daily activity is performed by human, its execution is adjusted according to the environment and is different from last time. Programming an industrial robot to accomplish the same is prohibitively difficult, whereas human completes those tasks with ease. To make robots more widely useful, researchers have been trying to help robots learn a task and generalize to different situations, to which the approach of teaching robots by providing examples has received considerable attention, known as programming by demonstration (PbD) \cite{billard2008}. In this work, we consider learning a task on the level of motion trajectories, and to perform a task, we generate a new trajectory.  

Pouring is a simple task that human practice daily. It is the second most frequently executed motion in cooking scenarios after pick-and-place \cite{paulius2016}. It relies on rotating a cup (or a container in general) that holds certain material. Sufficient rotation of the cup makes the material come out and sufficient recovery makes the pouring stop. Human typically pour using vision, while sensing the force resulted from the cup. We aim to learn from data how to pour using force feedback alone.               

One popular framework for motion trajectory generation is dynamical movement primitives (DMP) \cite{ijspeert_etal2013}. DMP is a stable non-linear dynamical system, and is capable of modeling discrete movement such as swinging a tennis racket \cite{ijspeert_etal2002}, playing table tennis  \cite{kober_etal2010} as well as rhythmic movement such as drumming \cite{schaal2003} and walking \cite{nakanishi2004}. DMP consists of a non-linear forcing function, a canonical system and a transformation system. The forcing function defines the desired task trajectory. The transformation system is a point attractor system for discrete movement or a limit cycle system for rhythmic movement that is modulated by the forcing function. The canonical system is a well-understood dynamical system that is guaranteed to converge and also serves to provide the phase variable of a task. The parameters that represent the forcing function can be learned from human demonstrations using locally weighted regression \cite{schaal1998} or other regression methods.

Based on DMP, \cite{amor_etal2014} introduced interactive primitives for two-agent collaborative tasks. A predictive distribution of the parameters of DMP is maintained and is used to infer the collaborative activity of one agent while observing that of the other. \cite{ewerton2015} extends \cite{amor_etal2014} by using a Gaussian mixture of interactive primitives. Such extension allows the correlation between two agents of a collaboration task to be non-linear, and it also enables modeling multiple collaborative tasks.       

Another approach for motion generation is based on Gaussian mixture model (GMM) and Gaussian mixture regression (GMR) \cite{calinon09book}. GMM is used to model the trajectories of a task and GMR is used for task reproduction. GMM is learned using all the variables of a movement including time stamps, and GMR is conducted by infering the movement variables using the learned GMM conditioned on the time stamp. The parameters of GMM can be learned using the Expectation-Maximization algorithm \cite{dempster1977}. Several constraints can be incorporated using the product property of Gaussians while performing a new task. The approach can be applied to both world and joint space to consider the trade-off of variations in difference spaces and thus achieves a more accurate control \cite{calinon2008IROS}. Also, task-parameterized GMM models a movement using multiple candidate frames of reference, and thus enables more detailed motion production \cite{calinon_etal2012}. Using time as the index the approach can produce a task trajectory in a single shot. In comparison, the approach can be extended to model a dynamical system which produces a trajectory step by step \cite{calinon2009}.

Principal Component Analysis (PCA) also proves useful for motion generation. Known as a dimension reduction technique used on the dimentionality axis of the data, PCA can be used on the time axis of motion trajectories instead to retrieve geometric variations \cite{lim2005}. Besides, PCA can also be applied to find variations in how the motion progresses in time, which, combined with the variations in geometry enables generating motions with more flexibility \cite{min2009}. Functional PCA (fPCA) extends PCA by introducing continuous-time basis functions and treating trajectories as functions instead of  collections of points \cite{ramsay_etal2009}. \cite{huang2015} applies fPCA for producing trajectories of gross motion such as answering phone and punching, and for making the trajectories avoid obstacles with the guidance of quality via points. \cite{paulius2016} uses fPCA for generating trajectories of fine motion such as pouring.      

Recently, recurrent neural networks (RNN) receives increasing attention. At any time step, RNN takes a given input and the output emitted from the last time step, and emits an output which is passed to the next time step. The mechanism of RNN makes it inherently suitable for handling sequential data. Similar to DMP \cite{ijspeert_etal2013} and GMR based approach \cite{calinon2009}, RNN is also capable of modeling general dynamical systems \cite{han2004, trischler2016}. RNN can be readily used to generate trajectories by relating the emitted output to future inputs. For example, \cite{graves2013} generates English hand writing trajectories by predicting the location offset of the tip of the pen and the end of a stroke. \cite{zhang_etal2016} applies a similar strategy to generate Chinese characters. \cite{fragkiadaki2015} generates motion capture trajectories by directly predicting the joint angle vector for the next time step.     

The paper goes as follows. In Section \ref{sec-approach}, we review the fundamentals of RNN and particularly LSTM, and present our pouring system. In Section \ref{sec-exp}, we describe the data collection and preparation process, training the system, and creating and training a separate force estimation system. In Section \ref{sec-res}, we conduct experiments to evaluate whether our system generalizes to unseen situations. We discuss the performance of our pouring system in Section \ref{sec-conc}.

\section{Methodology for Pouring Trajectory Generation} \label{sec-approach}

In this section, we describe in detail our system of generating a pouring trajectory which builds on long short-term memory. To explain why we choose RNN as the building block, prior to the system description, we review the basics of traditional RNN, and of one particular structure, the long short-term memory. 

\subsection{Recurrent Neural Network} \label{sec-rnn}
Recurrent neural network (RNN) conducts its computation one step at a time, and at any step its input consists of two parts: a given input, and its own output from the previous time step. The idea is shown in Eq. \eqref{eq-rnn} where $x_t$ is the given input, $h_{t-1}$ and $h_t$ are output from the previous and at the current step. The weight $W$ and bias $b$ can  be learned using Backpropagation Through Time \cite{werbos1990}.  
\begin{equation} \label{eq-rnn}
h_t = \text{tanh}\left(W[h_{t-1}, x_t]^\top + b\right)
\end{equation}

\begin{figure}
    \includegraphics[width=\linewidth]{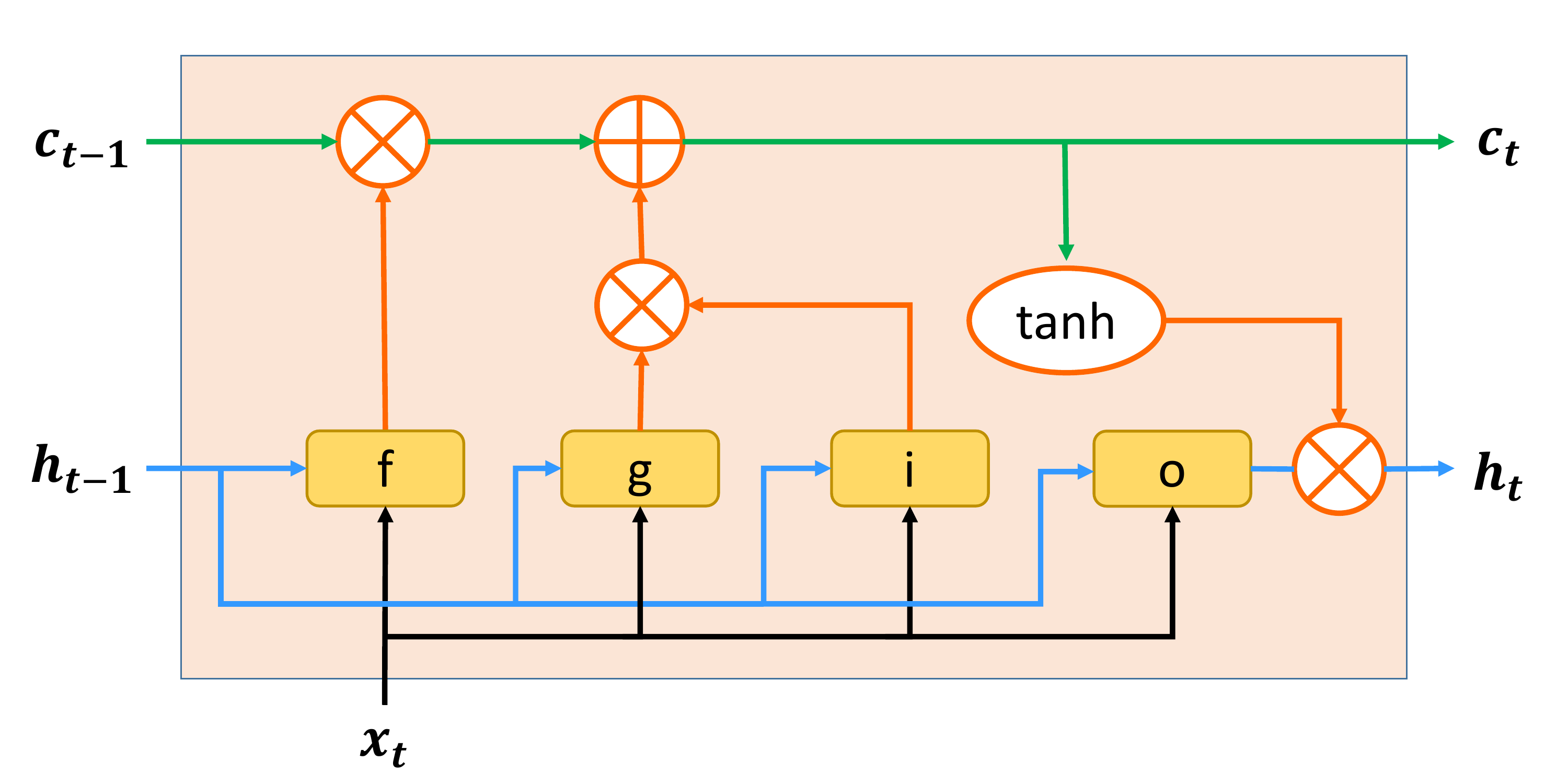}
    \caption{Mechanism inside an LSTM unit, Zaremba's version \cite{zaremba2014}}
    \label{fig-lstm}
\end{figure}

In theory, by including its past output in its input, RNN takes the entire history of given inputs into account when it conducts computation at any step, and therefore is inherently suitable for handling sequential data. However, the traditional RNN as shown in Eq. \eqref{eq-rnn} is difficult to train and has vanishing gradients problem, and therefore is inadequate for problems involving long-term dependency \cite{bengio1994, hochreiter1997}. Long short-term memory (LSTM) is a specific RNN design that overcomes the vanishing gradient problem \cite{hochreiter1997}. We use a version of LSTM whose working mechanism is described by \cite{zaremba2014}:
\begin{align}
    i &= \text{sigm}\left(W_i[h_{t-1}, x_t]^\top + b_i\right) \\
    o &= \text{sigm}\left(W_o[h_{t-1}, x_t]^\top + b_o\right) \\
    f &= \text{sigm}\left(W_f[h_{t-1}, x_t]^\top + b_f\right) \\
    g &= \text{tanh}\left(W_g[h_{t-1}, x_t]^\top + b_g\right) \\
    c_t &= f \odot c_{t-1} + i \odot g \\
    h_t &= o \odot \text{tanh}(c_t)
\end{align}        
where $i, o, f$ are the input, output, and forget gates respectively, $c$ is the cell, sigm is short for sigmoid, and $\odot$ represent element-wise multiplication. Fig. \ref{fig-lstm} gives an illustration.

LSTM has been proven successful for sequential generation applications including generating hand written characters \cite{graves2013, zhang_etal2016}, captioning images \cite{karpathy14} and videos \cite{venugopalan2015}, drawing images \cite{gregor2015}, translating natural languages \cite{sutskever2014, cho2014}, and executing computer programs \cite{zaremba14exe}.

We identify RNN, and specifically LSTM, as the architecture with which we build our pouring system. The reasons include:
\begin{enumerate}
\item The structure of RNN makes it inherently fit for handling sequences.
\item RNN is capable of modeling dynamical systems. Since a dynamical system is powered by velocity (or acceleration), it has the ability to react to changes of the environment.
\item RNN has proven ability to generate both categorical and continuous-valued sequences.
\item RNN eliminates the needs for temporally aligning sequences before modeling, and therefore preserves the dynamics in a sequence.
\item LSTM supercedes the traditional RNN, and has proven ability to handle long-term dependency. 
\end{enumerate} 

\subsection{Generating Pouring Trajectory}
The pouring system predicts the velocity of rotation using the force feedback produced by the cup, which is shown as (middle) in Fig. \ref{eq-rnn}.

\begin{figure}
    \includegraphics[trim={1.2cm 0.5cm 1.3cm 0.3cm},clip, width=\linewidth]{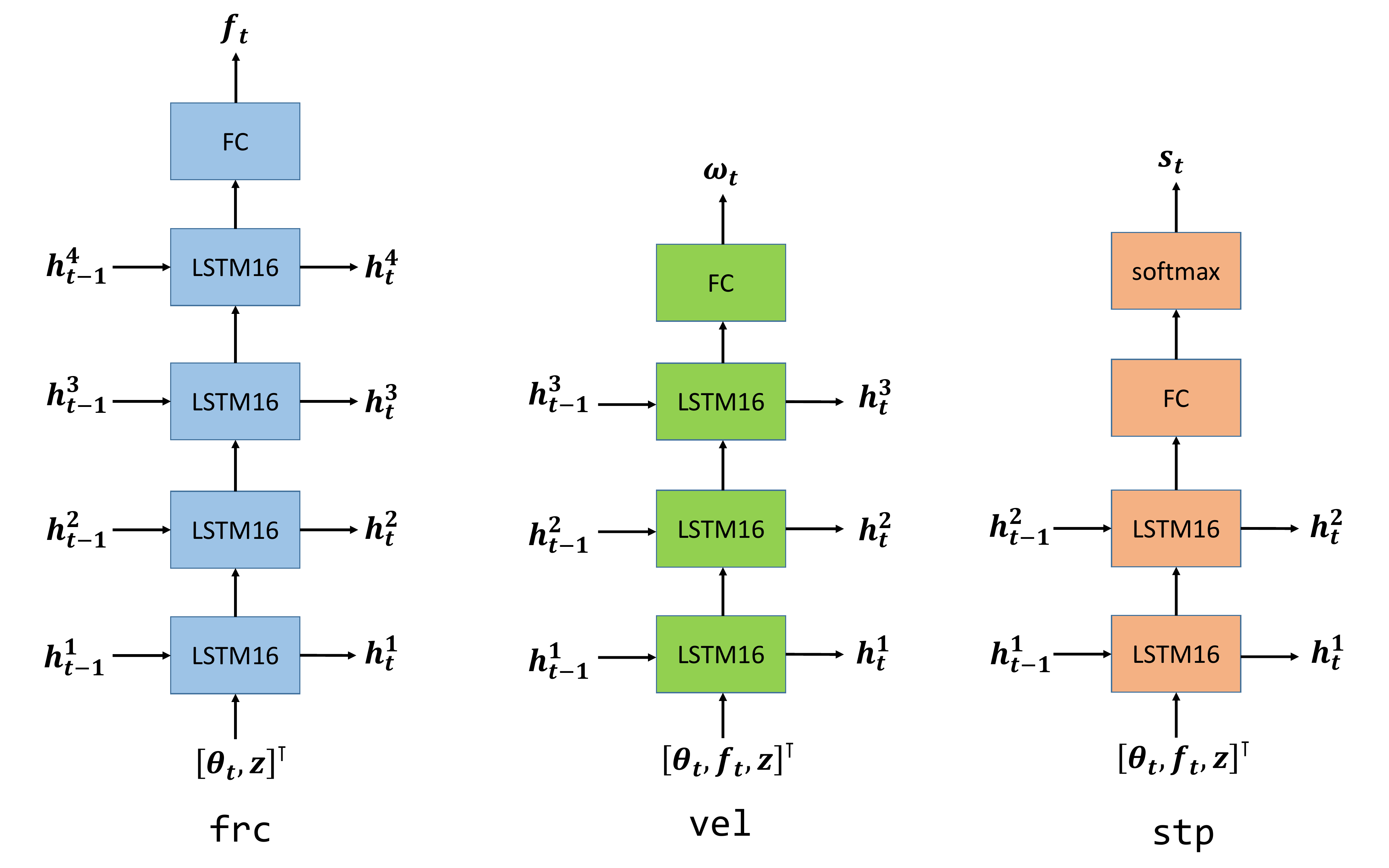}
    \caption{The architectures of (left) \texttt{frc}, (middle) \texttt{vel}, (right) \texttt{stp}. LSTM16 refers to 16 LSTM units. FC refers to fully connected.}
    \label{fig-system}
\end{figure}

We assume $n$ trials of pouring motion are available. The data of trial $i$ are represented by $( \theta_{1\dots T_i}, f_{1\dots T_i}, z)^{(i)}$, where $\theta_{1\dots T_i}$ is the sequence of cup rotation, $T_i$ is the sequence length, $f_{1\dots T_i}$ is the sequence of sensed force, and $z$ represents static data that characterize the trial. For simplicity, we assume $\theta, f, z$ are all one-dimensional. 

We refer to the system that predicts the velocity of rotation as \texttt{vel}. The actual velocity is computed by 
\begin{equation}
\omega_t = \theta_{t+1} - \theta_t, \quad t=1\dots T_i-1.
\end{equation}
At step $t$, \texttt{vel} takes $[\theta_t, f_t, z]^\top$ as input, and generates predicted velocity $\hat{\omega}_t$:
\begin{align}
    \textbf{h}_t &= \text{LSTM}([\theta_t, f_t, z]^\top)\\
    \hat{\omega}_t &= \text{fc}(\textbf{h}_t)
\end{align}
where `fc' is short for `fully connected'. The loss is defined using Euclidean distance: 
\begin{equation}
L_\text{vel} = \frac{1}{n}\sum_{i=1}^n\frac{1}{T_i-1}\sum_{t=1}^{T_i-1}(\omega_t^{(i)} - \hat{\omega}_t^{(i)})^2.
\end{equation}

%The trained \texttt{frc} can be used to generate more data, which can be used to train \texttt{vel}.

In order to automatically stop the generation process after the pouring task has completed, we create a stopping system. We refer to the system that stops the pouring motion as \texttt{stp} shown as (right) in Fig \ref{fig-system}, which is a binary classifier. At step $t$, \texttt{stp} takes $[\theta_t, f_t, z]$ as input, and outputs a 2-vector $\mathbf{r}_t$. We define class 0 as `continue', and class 1 as `stop'. 
\begin{align}
    \textbf{h}_t &= \text{LSTM}([\theta_t, f_t, z]^\top) \\
    \mathbf{r}_t &= \text{fc}(\textbf{h}_t) \\
    \mathbf{s}_t &= \text{softmax}(\mathbf{r}_t)    
\end{align}   
Let the target be represented by a trivial one-hot vector $\mathbf{s}'_t = [s'_{t, 1}, s'_{t, 2}]^\top$, where $s'_{t, 1}, s'_{t, 1} \in \{0, 1\}$ and $s'_{t, 1} + s'_{t, 2} = 1$. The loss is defined using cross entropy:
\begin{equation}
    L_\text{stp} = -\sum_{i=1}^n\sum_{t=1}^{T_i}\left(s'^{(i)}_{t, 1}\text{ln}s^{(i)}_{t, 1} +  s'^{(i)}_{t, 2}\text{ln}s^{(i)}_{t, 2}\right)
\end{equation}

\begin{figure}
    \begin{center}
    \includegraphics[width=0.8\linewidth]{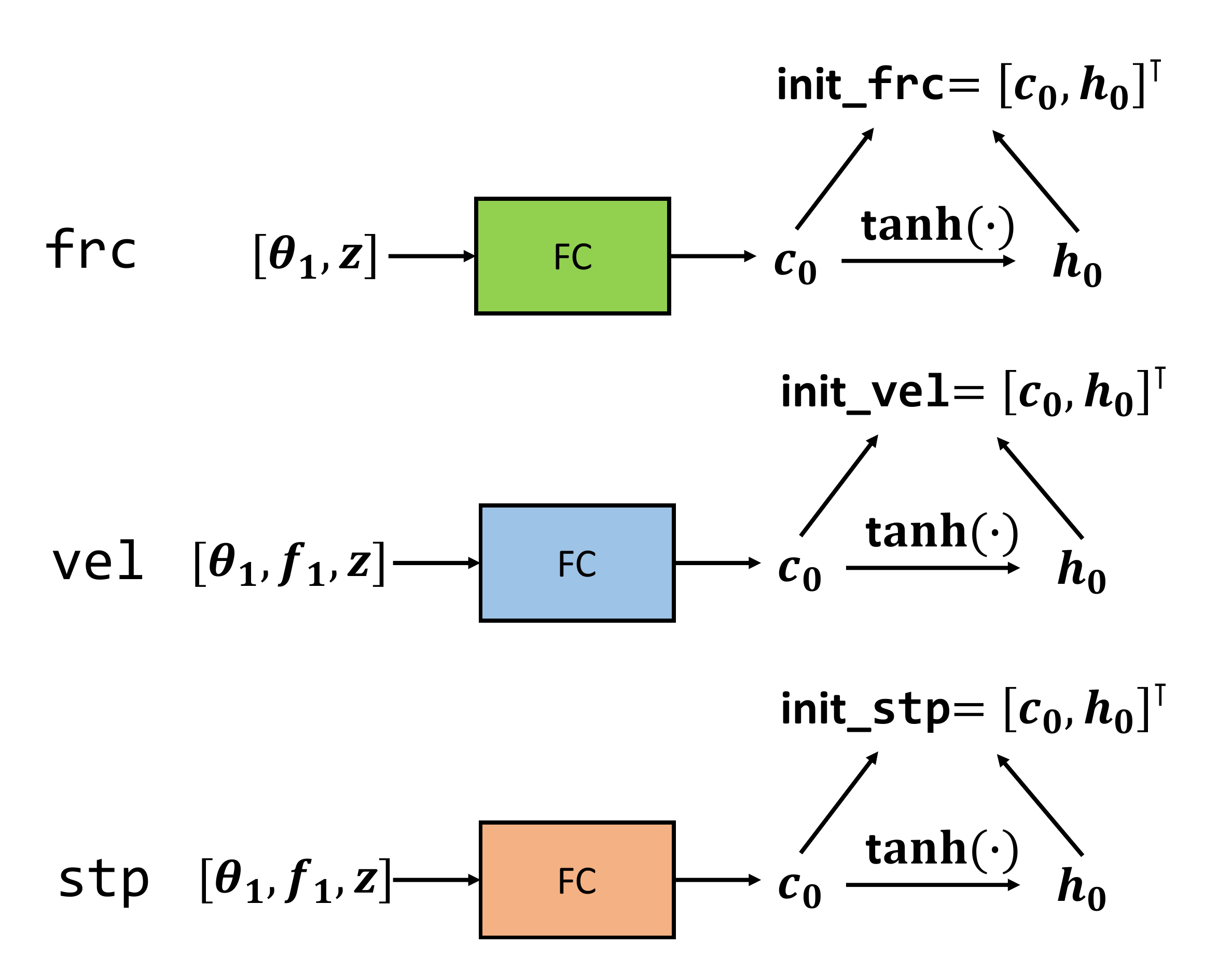}
    \caption{Initializing \texttt{frc}, \texttt{vel}, \texttt{stp}}
    \label{fig-init}
    \end{center}
\end{figure}

The initial state of LSTM includes $c_0$ and $h_0$, which are obtained by
\begin{align}
c_0 &= \text{fc}([\theta_1, f_1, z]^\top),\\
h_0 &= \text{tanh}(c_0),
\end{align}
as shown in Fig. \ref{fig-init}.

%Compared with setting $c_0 = h_0 = 0$, the initialization method above helps estimating the first force $f_1$ and predicting the first velocity $\omega_1$. 

%Fig.  illustrates the initialization. 

The trajectory is generated by first initializing \texttt{vel} and \texttt{stp}, and then keep generating and executing rotational velocities. Specifically, the trajectory generation process is described in Alg. \ref{alg-motion_gen}. 

\begin{algorithm}
    \caption{Trajectory Generation} % placed before \label
    \label{alg-motion_gen}
    \begin{algorithmic}[1]
        %\State Initialize \texttt{frc} using $[\theta_1, z]^\top$
        %\State $f_1 \gets \texttt{frc}([\theta_1, z]^\top)$
        \State Initialize \texttt{vel} and \texttt{stp} using $[\theta_1, f_1, z]^\top$
        \State $t \gets 1$
        \While {True}
        %\State $f_t \gets \texttt{frc}([\theta_t, z]^\top)$ 
        \State $\omega_t \gets \texttt{vel}([\theta_t, f_t, z]^\top)$
        \State $\theta_{t+1} \gets \theta_t + \omega_t$
        \State $s \gets \text{argmax }\texttt{stp}([\theta_t, f_t, z]^\top)$
        %\State $f_{t+1} \gets \texttt{frc}([\theta_{t+1}, z]^\top)$ 
        \State $t \gets t+1$
        \If {$s == 1$} 
        \State Break
        \EndIf
        \EndWhile
    \end{algorithmic}
\end{algorithm}

\section{Data Preparation and Training} \label{sec-exp}

\begin{figure}
    \includegraphics[width=\linewidth]{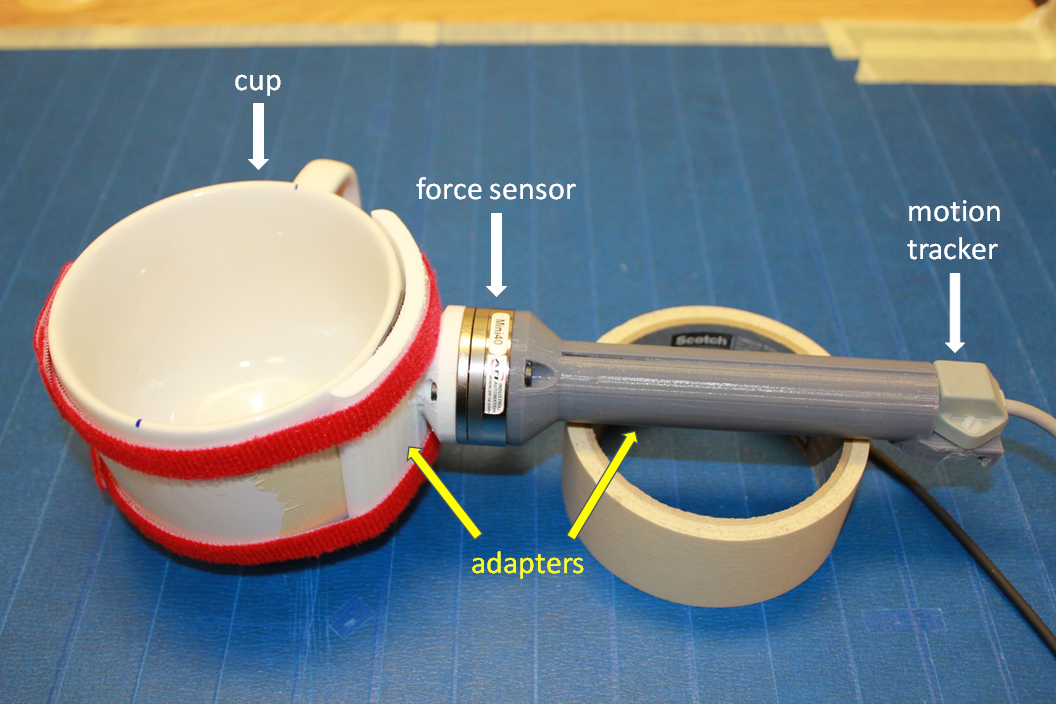}
    \caption{3D printed adapters that connect the cup, the force sensor, and the motion tracker.}
    \label{fig-adapter}
\end{figure}

The equipment for data collection includes six different cups, ten different containers, one ATI mini40 force and torque (FT) sensor, and one Polhemus Patriot motion tracker. We refer to the pour-from container as \emph{cup} and the pour-to container as \emph{container}. All cups are mutually different and so are all the containers. The FT sensor records $(f_x, f_y, f_z, \tau_x, \tau_y, \tau_z)$ at 1KHz. The motion tracker records $(x, y, z, \text{yaw}, \text{pitch}, \text{roll})$ at 60Hz. The cup, the force sensor, and the motion tracker are connected by 3D printed adapters, shown in Fig. \ref{fig-adapter}. The materials that are poured include water, beans, and ice. 

We obtain the \emph{empty} reading by keeping an empty cup in a level position, taking 500 FT samples (which takes 0.5 second), and then taking the average. Similarly, for each trial, we obtain the \emph{initial} reading right before the trial with material in the cup, and the \emph{final} reading right after the trial with or without material in the cup depending on the trial.

We define the sensed force as 
\begin{equation}
f = \sqrt{f_x^2 + f_y^2 + f_z^2}.
\end{equation}

In total we collected 1,138 trials which involves 3 subjects. Each trial is represented by a sequence $\{\mathbf{a}_t\}_{t=1}^{T_i}$ where $\mathbf{a}_t\in\mathbb{R}^{10}$ and 

\begin{tabular}{@{}p{0.4cm} p{0.8cm} p{6cm}@{}}
$\mathbf{a}_t=[$&\\
&$\theta_t$ & rotation angle at time $t$ (degree)\\
&$f_t$ & sensed force at time $t$ (lbf)\\
&$f_\text{init}$  & sensed force before pouring (lbf)  \\ 
&$f_\text{empty}$ & sensed force while cup is empty (lbf) \\
&$f_\text{final}$ & sensed force after pouring (lbf)\\
&$d_\text{cup}$ & diameter of the cup (mm)\\
&$h_\text{cup}$ & height of the cup (mm)\\
&$d_\text{ctn}$ & diameter of the container (mm)\\
&$h_\text{ctn}$ & height of the container (mm)\\
&$\rho$ & material density / water density (unitless)\\
$]$
\end{tabular}

We pad all the sequences to the maximum length in the data: $T_{max} = \text{max}(\{T_i\})$. For \texttt{vel}, we pad using zero because zero padding makes it easy to compute the original length of a sequence during training. For \texttt{stp}, we pad using the end value of the sequence because \texttt{stp} is intended to be used on generated motions which will not have zero padding.  

We train using the Adam optimizer \cite{kingma2014} and set the learning rate to 0.01. We trained each system for a fixed number of epochs: 4,000 for \texttt{vel}, and 2,000 for \texttt{stp}. The training error for \texttt{vel} ranges from 0.002 to 0.005 (mm), and the accuracy of \texttt{stp} ranges from 0.9 to 0.98. 

\subsection{Training Force Estimation}
In order to run our approach in simulation, we need to have force feedback after we have arrived at a new rotation. Real force feedback is not applicable in simulation. The movement of the liquid during pouring forms a complex dynamical system and is difficult to calculate analytically. Thus, to get force feedback, we decide to generate the force by ourselves. To that end, we learn from data the mapping relationship from rotation angles to force, and then use the learned model to estimate the force corresponding to current rotation.

Thus, we need to train a new system. We refer to the system that estimates the sensed force from rotation as \texttt{frc}, shown as (left) in Fig. \ref{fig-system}. At step $t$, \texttt{frc} takes $[\theta_t, z]^\top$ as input, and produces estimated force $\hat{f}_t$:
\begin{align}
\textbf{h}_t &= \text{LSTM}([\theta_t, z]^\top) \\
\hat{f}_t &= \text{fc}(\textbf{h}_t)
\end{align}

The loss is defined using Euclidean distance:
\begin{equation}
L_\text{frc} = \frac{1}{n}\sum_{i=1}^n\frac{1}{T_i}\sum_{t=1}^{T_i}(f_t^{(i)} - \hat{f}_t^{(i)})^2.
\end{equation}

The initialization of \texttt{frc} includes
\begin{align}
c_0 &= \text{fc}([\theta_1, z]^\top), \\ 
h_0 &= \text{tanh}(c_0),
\end{align}
as shown in Fig. \ref{fig-init}.
 
The data preparation for \texttt{frc} uses zero padding. We train the \texttt{frc} with a fixed 2000 epochs, and the error ranges between 0.002 to 0.003 (lbf).  

With \texttt{frc}, the trajectory generation process needs modification. Force can no longer be assumed to be available, but must be produced explicitly by \texttt{frc}. The modified trajectory generation process is shown in Alg. \ref{alg-motion_gen_sim}. 

\begin{algorithm}
    \caption{Trajectory generation for simulation} % placed before \label
    \label{alg-motion_gen_sim}
    \begin{algorithmic}[1]
        \State Initialize \texttt{frc} using $[\theta_1, z]^\top$
        \State $f_1 \gets \texttt{frc}([\theta_1, z]^\top)$
        \State Initialize \texttt{vel} and \texttt{stp} using $[\theta_1, f_1, z]^\top$
        \State $t \gets 1$
        \While {$t < T_{max}$}
        %\State $f_t \gets \texttt{frc}([\theta_t, z]^\top)$ 
        \State $\omega_t \gets \texttt{vel}([\theta_t, f_t, z]^\top)$
        \State $\theta_{t+1} \gets \theta_t + \omega_t$
        \State $s \gets \text{argmax }\texttt{stp}([\theta_t, f_t, z]^\top)$
        \State $f_{t+1} \gets \texttt{frc}([\theta_{t+1}, z]^\top)$ 
        \State $t \gets t+1$
        \If {$s == 1$} 
        \State Break
        \EndIf
        \EndWhile
    \end{algorithmic}
\end{algorithm}

\section{Experiment on Generalization} \label{sec-res}

We evaluate the generalization ability of our approach and see if it can generate pouring motion in unseen situations. Given a test sequence, we extract $\theta_1$ and $z$, and generate a sequence using Alg. \ref{alg-motion_gen_sim}. The evaluation is conducted in simulation.

We test the system using unseen 
\begin{enumerate}
    \item cup,
    \item container,
    \item material,
    \item cup and container,
    \item container and material,
    \item cup and material,
    \item cup and container and material.
\end{enumerate}

\subsection{Identifying success}
We evaluate the generalization ability of the pouring system using dynamic time warping (DTW) \cite{sakoe1978}, which gives the minimum normalized distance between two trajectories. 

We provide a set of test sequence which include an element that is unseen during training and see if the system is able to adapt to the changes. Let the set of test sequences be $\{x_i\}_{i=1}^{m}$. We first compute the distance between each pair of test sequences and draw a histogram:
\begin{equation}
h_1 = \text{hist}(\{\text{dtw}(x_i, x_j)\}_{i\neq j}) \quad i, j =1, 2, \dots, m.
\end{equation} 
Each $x_i$ can be used to generate a new trajectory $x_i'$. We compute the distance between $x_i'$ and every test sequence $x_j$ and draw another histogram. 
\begin{equation}
h_2 = \text{hist}(\{\text{dtw}(x_i', x_j)\}) \quad i, j =1, 2, \dots, m.
\end{equation}
Both histograms are normalized. We visually compare the similarity between $h_1$ and $h_2$. If they are similar, then it means the generated trajectories are similar to the trajectories executed by human, which identifies that the generalization succeeds. The system fails to generalize if otherwise.

\subsection{Results}

\begin{figure}
    \includegraphics[width=\linewidth]{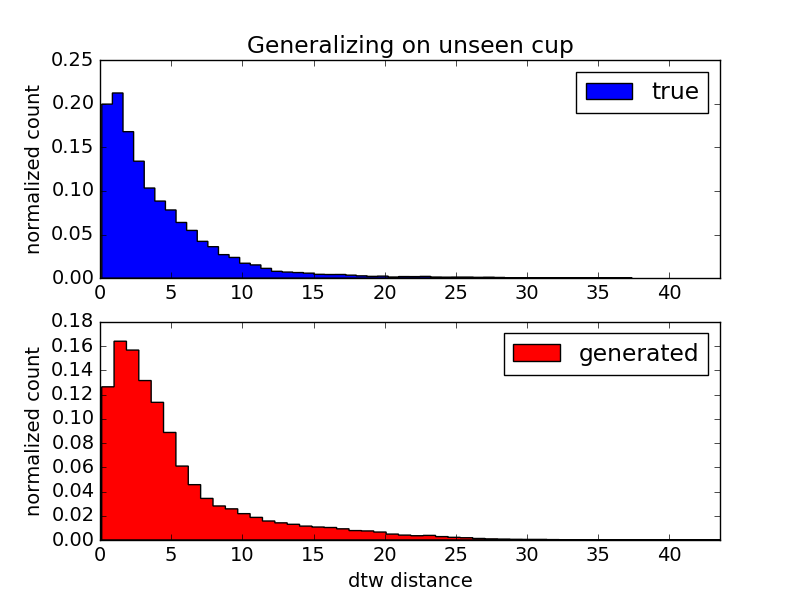}
    \caption{Generalizing on an unseen cup.}
    \label{fig-cup}
\end{figure}

\begin{figure}
    \includegraphics[width=\linewidth]{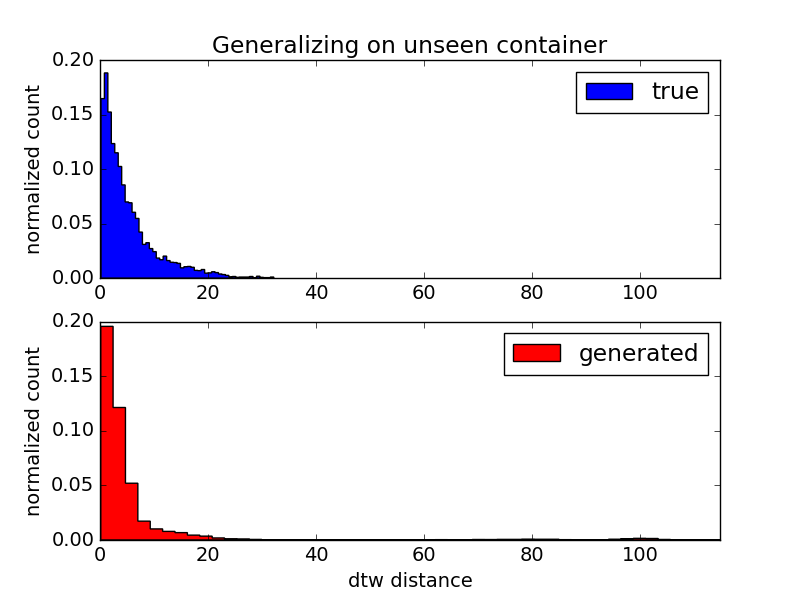}
    \caption{Generalizing on an unseen container}
    \label{fig-ctn}
\end{figure}

\begin{figure}
    \includegraphics[width=\linewidth]{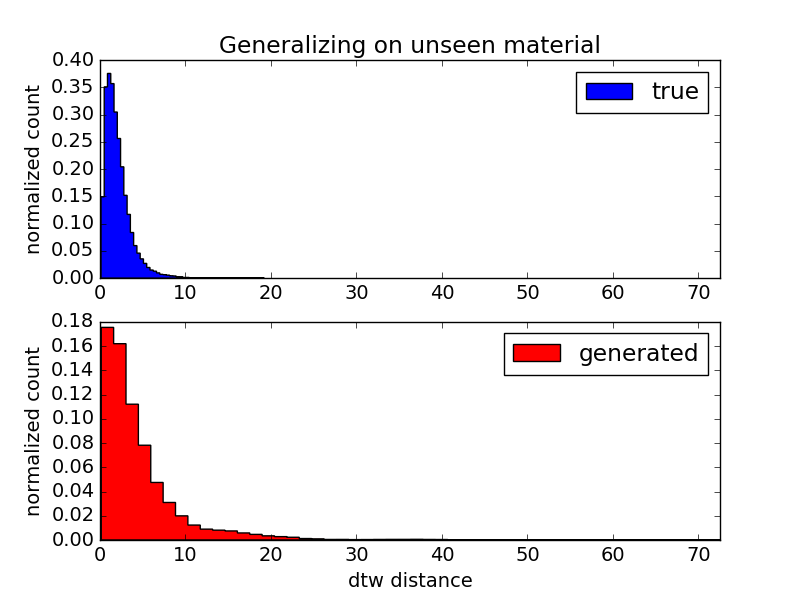}
    \caption{Generalizing on an unseen material}
    \label{fig-mat}
\end{figure}

\begin{figure}
    \includegraphics[width=\linewidth]{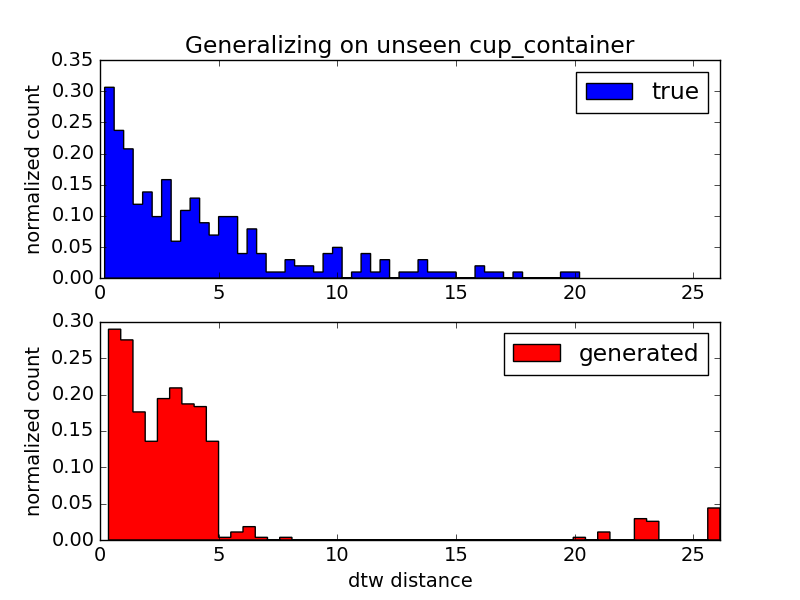}
    \caption{Generalizing on an unseen cup and container}
    \label{fig-cup_ctn}
\end{figure}

\begin{figure}
    \includegraphics[width=\linewidth]{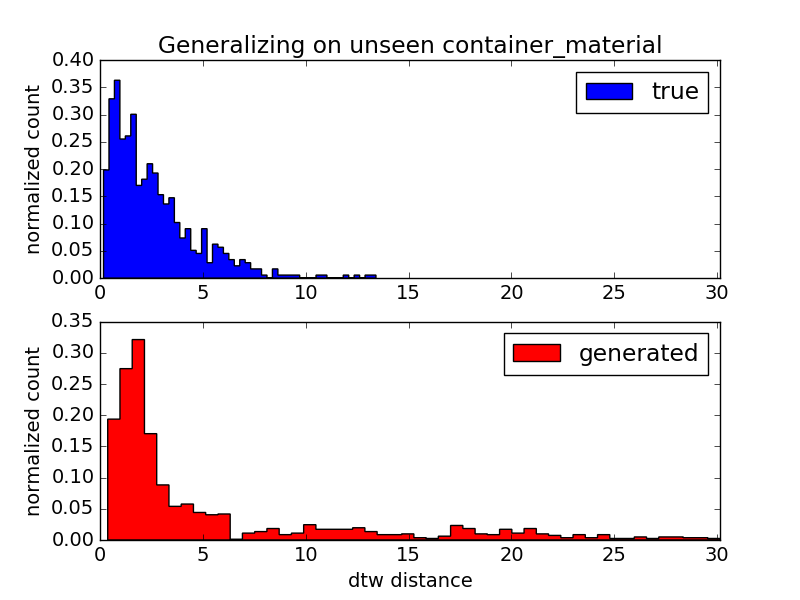}
    \caption{Generalizing on an unseen container and material}
    \label{fig-ctn_mat}
\end{figure}

\begin{figure}
    \includegraphics[width=\linewidth]{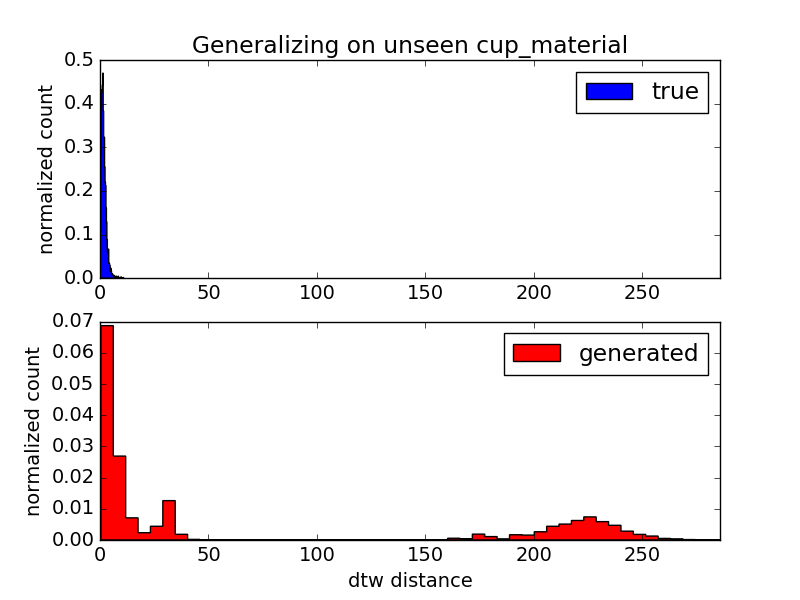}
    \caption{Generalizing on an unseen cup and material}
    \label{fig-cup_mat}
\end{figure}

\begin{figure}
    \includegraphics[width=\linewidth]{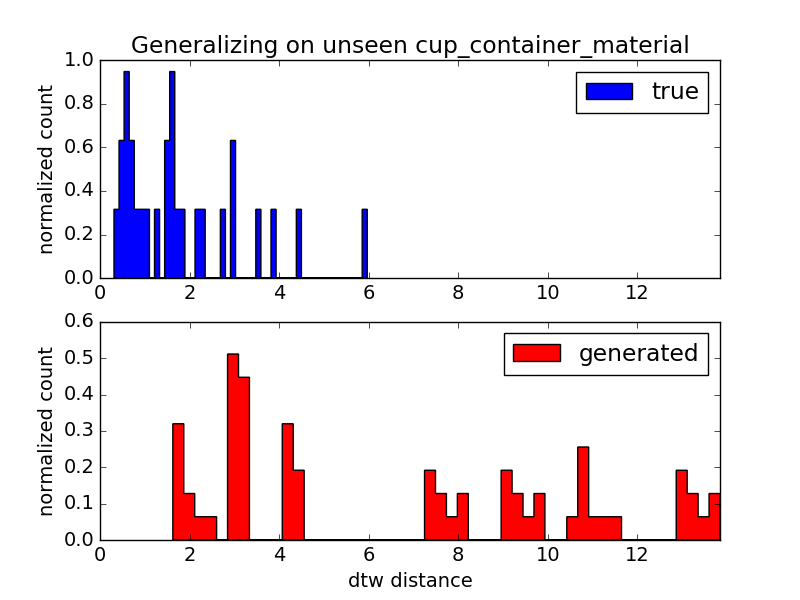}
    \caption{Generalizing on an unseen cup, container, and material}
    \label{fig-cup_ctn_mat}
\end{figure}

The results for the seven cases of unseen elements of the pouring characteristics are shown in Fig. \ref{fig-cup} to \ref{fig-cup_ctn_mat}. Generalization on cup, or container, or material alone is successful because the pairing histograms are similar (Fig. \ref{fig-cup}, \ref{fig-ctn} and \ref{fig-mat}). Generalizing on cup and container (Fig. \ref{fig-cup_ctn}) and container and material (Fig. \ref{fig-ctn_mat}) can be considered successful because of the similarity in the concentration of the small-distances, despite the difference on mid to high-valued distance parts, which occupy only a small portion of all the distances. Generalizing on cup and material fails as well as on cup and container and materials, as shown in Fig. \ref{fig-cup_mat} and Fig. \ref{fig-cup_ctn_mat}. For cup and container and material, only 8 test sequences are available, which may partly contribute to the difference between the two histograms.   
 
\section{Discussion} \label{sec-conc}
We have presented an approach of generating pouring trajectories by learning from pouring motions demonstrated by human subjects. The approach uses force feedback from the cup to determine the future velocity of pouring. We aim to make the system generalize its learned knowledge to unseen situations. The system successfully generalize when either a cup, a container, or the material changes, and starts to stumble when changes of more than one element are present. Since the total size of data does not change, the more that is left out for testing (more unseen elements), the less there is available for training. Thus, the system accepts weaker training and after which faces more demanding challenges. The observed results of degrading performance with increading generalization difficulty is expected.

We have started evaluating the system on an industrial robot that is equipped with a force sensor. The evaluation is still under way. 

Future work includes finishing the evaluation on the industrial robot, designing a quantitative measure that measures the degree of success of a generated trajectory, modifying the architecture to emphasize the role of initial and final force, and getting help from reinforcement learning.

%\section*{Acknowledgments}
%Thank NSF?

\bibliographystyle{IEEEtran}
\bibliography{rss_ref}

\end{document}